%% file: main.tex
\documentclass[10pt, a4paper]{article}
\usepackage{lrec2022} 
\usepackage{multibib}
\newcites{languageresource}{Language Resources}
\usepackage{graphicx}
\usepackage{tabularx}
\usepackage{soul}
\usepackage{xcolor}
\usepackage{titlesec}
\titleformat{\section}{\normalfont\large\bfseries\center}{\thesection.}{1em}{}
\titleformat{\subsection}{\normalfont\SmallTitleFont\bfseries\raggedright}{\thesubsection.}{1em}{}
\titleformat{\subsubsection}{\normalfont\normalsize\bfseries\raggedright}{\thesubsubsection.}{1em}{}
\renewcommand\thesection{\arabic{section}}
\renewcommand\thesubsection{\thesection.\arabic{subsection}}
\renewcommand\thesubsubsection{\thesubsection.\arabic{subsubsection}}

\usepackage{epstopdf}
\usepackage[utf8]{inputenc}

\usepackage{hyperref}
\usepackage{xstring}

\usepackage{color}

\title{Hierarchical Annotation for Building A Suite of Clinical Natural Language Processing Tasks: Progress Note Understanding}

\name{Yanjun Gao$^1$, Dmitriy Dligach$^2$, Timothy Miller$^3$, Samuel Tesch$^4$, Ryan Laffin$^4$, \\
\bf \large  Matthew M. Churpek$^1$, Majid Afshar$^1$} 

\address{$^1$ICU Data Science Lab, School of Medicine and Public Health, 
    University of Wisconsin \\ 
    $^2$Loyola University Chicago \\
         $^3$Boston Children's Hospital and Harvard Medical School  \\
         $^4$School of Medicine and Public Health, 
    University of Wisconsin \\ 
        $^1$\{ygao, mchurpek, mafshar\}\@medicine.wisc.edu \\
         $^2$ddligach@luc.edu\\
         $^3$Timothy.Miller@childrens.harvard.edu \\
         $^4$\{sgtesch, rlaffin\}@wisc.edu
         }

\abstract{
Applying methods in natural language processing on electronic health records (EHR) data is a growing field. Existing corpus and annotation focus on modeling textual features and relation prediction. However, there is a paucity of annotated corpus built to model clinical diagnostic thinking, a process involving text understanding, domain knowledge abstraction and reasoning. This work introduces a hierarchical annotation schema with three stages to address clinical text understanding, clinical reasoning, and summarization. We created an annotated corpus based on an extensive collection of publicly available daily progress notes, a type of EHR documentation that is collected in time series in a problem-oriented format. The conventional format for a progress note follows a Subjective, Objective, Assessment and Plan heading (SOAP). We also define a new suite of tasks, Progress Note Understanding, with three tasks utilizing the three annotation stages. The novel suite of tasks was designed to train and evaluate future NLP models for clinical text understanding, clinical knowledge representation, inference, and summarization. 
 \\ \newline  }

\begin{document}

\maketitleabstract

\section{Introduction}
\input{introduction}

\section{Background}
\input{background}

\section{Data and Preparation}
\input{data}

\section{Annotation Protocol}
\input{annotation}

\section{Annotation Results}
\input{results}

\section{Suite of Tasks for Progress Note Understanding}

\input{task1}

\section{Conclusions}
We introduce a novel and hierarchical annotation on a large collection of publicly available EHR data and aim to develop and evaluate models for automated section segmentation, assessment and plan reasoning, and diagnoses summarization. A suite of tasks, Progress Note Understanding, is proposed to utilize the annotation and contains three tasks, each of which corresponds to an annotation stage. Future work will focus on hosting shared tasks for the clinical NLP community and using the tasks to build systems for applications in clinical decision support.  

\paragraph{Acknowledgement} We thank Professor Ozlem Uzuner (George Mason University) and Harvard Department of Biomedical Informatics (DBMI) for the feedback and guidance during N2C2 task setup. 
\section{Reference}
\label{lr:ref}
\bibliographystyle{lrec2022-bib}
\bibliography{references}

\end{document}

%% file: introduction.tex

Patients in the hospital have a multidisciplinary team of physicians, nurses, and support staff who attend to their care. As part of this care, providers input \textbf{daily progress notes} to update the diagnoses and treatment plan, and to document changes in the patient’s health status. The electronic health record (EHR) contains these daily progress notes, and they are one of the most frequent note types that carry the most relevant and viewed documentation of a patient's care~\cite{brown2014physicians}. While EHRs are intended to provide efficient care, they are still riddled with problems of note bloat (copying and pasting), information overload (automatically embedded data and administrative documentation), and poorly organized notes that overwhelm physicians and lead to burnout and, ultimately, inefficient care~\cite{shoolin2013association}.  

Applying methods in natural language processing to the EHR is a growing field with many potential applications in clinical decision support and augmented care. Corpus and annotation on EHR data are created to model semantic features and relation through linguistic cues, 
including relation extraction~\cite{mowery2008temporal}, named entity recognition~\cite{wang2009annotating,patel2018annotation,lybarger2021annotating}, question answering~\cite{pampari2018emrqa,raghavan2021emrkbqa}, natural language inference~\cite{romanov2018lessons}, etc. However, few corpora have been built to model clinical thinking, especially about \textit{clinical diagnostic reasoning}, a process involving clinical evidence acquisition, generating hypothesis, integration and abstraction over medical knowledge and synthesizing a conclusion in the form of a diagnosis and treatment plan~\cite{bowen2006educational}. 
In this work, we introduce a hierarchical annotation with three stages addressing clinical text understanding, reasoning and abstraction over evidence, and diagnosis summarization. The annotation guidelines were designed and developed by physicians with clinical informatics expertise in conjunction with computational linguistic experts to model the healthcare provider decision making process. Our annotations were built on top of the Medical Information Mart for Intensive Care-III (MIMIC-III), a publicly available English-language EHR~\cite{johnson2019mimic}.\footnote{MIMIC is available at \url{https://physionet.org/content/mimiciii/1.4/}. Data usage agreement is required. Our annotation will be provided through note ID, character offsets and labels upon acceptance. } 

Meanwhile, there exists only a handful of language benchmark tasks in medicine such as the National Library of Medicine BLUE Benchmark~\cite{peng2019transfer}, Biomedical Language Understanding and Reasoning Benchmark~\cite{gu2021domain}, and shared tasks hosted through Harvard’s National NLP Clinical Challenges (N2C2).\footnote{https://n2c2.dbmi.hms.harvard.edu/}. Some of these benchmarks contain non-EHR data~\cite{peng2019transfer,gu2021domain}. Many of the EHR-based benchmarks are time-insensitive, such as discharge summaries~\cite{mullenbach-etal-2021-clip,uzuner2008identifying}, radiology reports~\cite{abacha2021overview,peng2018negbio}. They also have a strong focus on modeling clinical language instead of potentials for clinical applications with a practitioner-derived focus~\cite{hultman2019challenges}. Recent advances in large scale language modeling enables pre-training on massive corpora and fine-tuning for in-domain tasks, such as transfer learning for BERT~\cite{devlin2019bert,he2020infusing} with ClinicalBERT~\cite{alsentzer2019publicly,hao2020enhancing}. These pre-trained models are evaluated on existing benchmarks with a great emphasis on domain knowledge representation, but few are tested for clinical reasoning and inference. Better benchmarks for clinical use are needed to evaluate clinical NLP systems that model clinical reasoning and thinking.  

The goal of the proposed research is to build a suite of tasks for the following: (1) identify the relevant sections of the daily progress note - Subjective, Objective, Assessment and Plan format (\textit{SOAP Note}); (2) identify problems/diagnoses from a set of daily progress notes; and (3) accurately discriminate related plans for an assessment in the Assessment and Plan section (A/P of the SOAP note) of daily progress notes. We refer to our suite of tasks as \textit{Progress Note Understanding}, which covers automatic section segmentation, Assessment and Plan reasoning, and diagnoses summarization. The three tasks are built from different levels of text units and annotation stages.  

\begin{figure}
    \centering
    \includegraphics[scale=0.4]{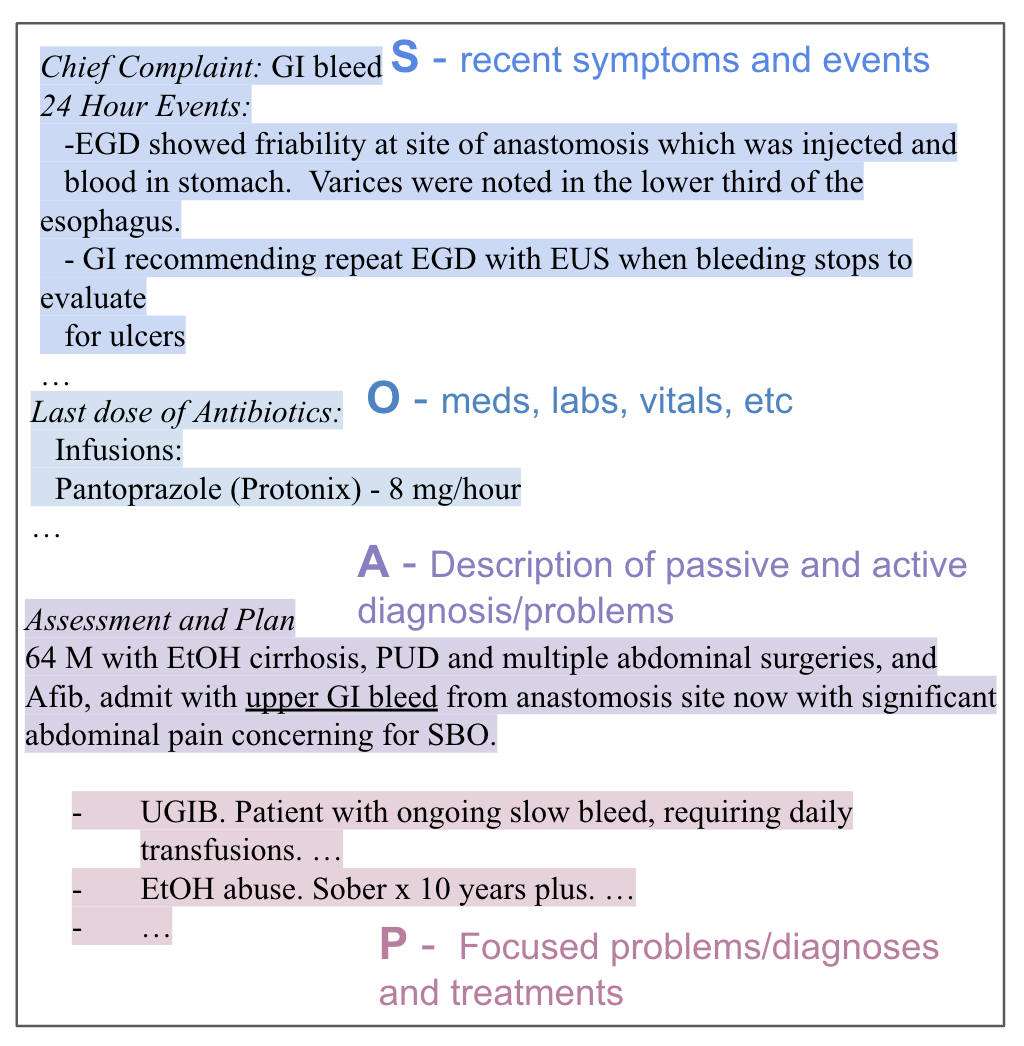}
    \vspace{-0.8pc}
    \caption{An example progress note with SOAP sections annotated. Line breaks are preserved from the raw note.}
    \label{fig:example_note}
\end{figure}

Our contribution includes: (1) a novel and hierarchical annotation protocol jointly designed by NLP researchers and physicians; (2) annotated data for training and evaluating NLP models on clinical text understanding, reasoning and text generation; (3) a new suite of tasks proposed to develop and evaluate NLP models for clinical applications that help to improve efficiency of bedside care and reduce medical errors. 


%% file: background.tex

The daily progress note follows a specific format with four major components: Subjective, Objective, Assessment, and Plan (SOAP). SOAP note documentation is engrained in medical school curriculum as well as other training curricula, developed by Larry Weed, MD, known as the father of the problem-oriented medical record and inventor of the ubiquitous SOAP daily progress note~\cite{weed1964medical}. It provides an easily recognizable template and pattern for systematic documentation, which lends itself also to reliable annotation. The main purpose of SOAP documentation is to record the patients' information, including recent events in their care and active problems in a readable and structured way so the patients' diagnoses are readily identified. This is separate from other note types, such as the admission note (history and physical, also known as H\&P) or discharge summary, which are not documented daily. The progress note is a time-series data collection with a reproducible format on all hospitalized patients, and it is the most frequently viewed note by care teams~\cite{hripcsak2011use}. As a patient's condition becomes more severe with an increasing number of interactions with providers, the progress note may also increase in length.   

Figure~\ref{fig:example_note} shows an example of a daily progress note with the sections and their corresponding subsections labeled. Every subsection in the progress note belongs to a section of SOAP.  Subjective includes sections of free text describing patients' symptoms, conditions, daily changes in care, etc. Objective contains sections of structured data such as lab results, vital sign, radiology reports. Assessment and Plan sections are considered by providers as the most important components in SOAP note, synthesizing evidences from Subjective and Objective and concluding the diagnoses and treatment plans. Specifically, Assessment is the section describing the patient and establishing the main symptoms or problems for their encounter, and Plan addresses each differential diagnosis/problem with an action plan or treatment plan for the day, so called \textsc{Plan Subsection}.  Figure~\ref{fig:example_note} contains two \textsc{Plan Subsection}, marked by purple color. Text \textit{``UGIB $\ldots$ daily transfusions"} is a different \textsc{Plan Subsection} than \textit{``EtOH abuse $\ldots$ to years plus"}. Each \textsc{Plan Subsection} starts with a summary of a problem/diagnosis, e.g. \textit{UGIB} in the second subsection summarizing from \textit{admit with upper GI bleed} in the Assessment section. 

The section headings are not always available in the raw progress notes. Depending on the patients' conditions, some sections are not always required to be present; hence, the number of sections in the progress note may vary. In the end, the SOAP note is reflective of the provider's effort to collect the most recent and relevant data and synthesize the collected information into a coherent understanding of patient's condition for decision-making and to ensure coordination of care. This skill in documentation requires clear reasoning to link symptoms, lab and imaging results, and other observations into temporally relevant and problem-specific treatment plans.



%% file: data.tex

\subsection{Annotation Tool and Data Source}

All annotation procedures were performed using a dedicated software tool called INCEpTION\footnote{v0.19.2 (Released on 2021-04-07), available at: \url{ https://inception-project.github.io/}}~\cite{klie2018inception}. INCEpTION is an open-source software tool that serves as a semantic annotation platform and offers intelligent assistance and knowledge management.  A sampling of 5,000 unique daily progress notes using the method described below were imported into the INCEpTION data corpus from the MIMIC-III notes file.

\begin{figure}[h]
    \centering
    \includegraphics[width=\columnwidth]{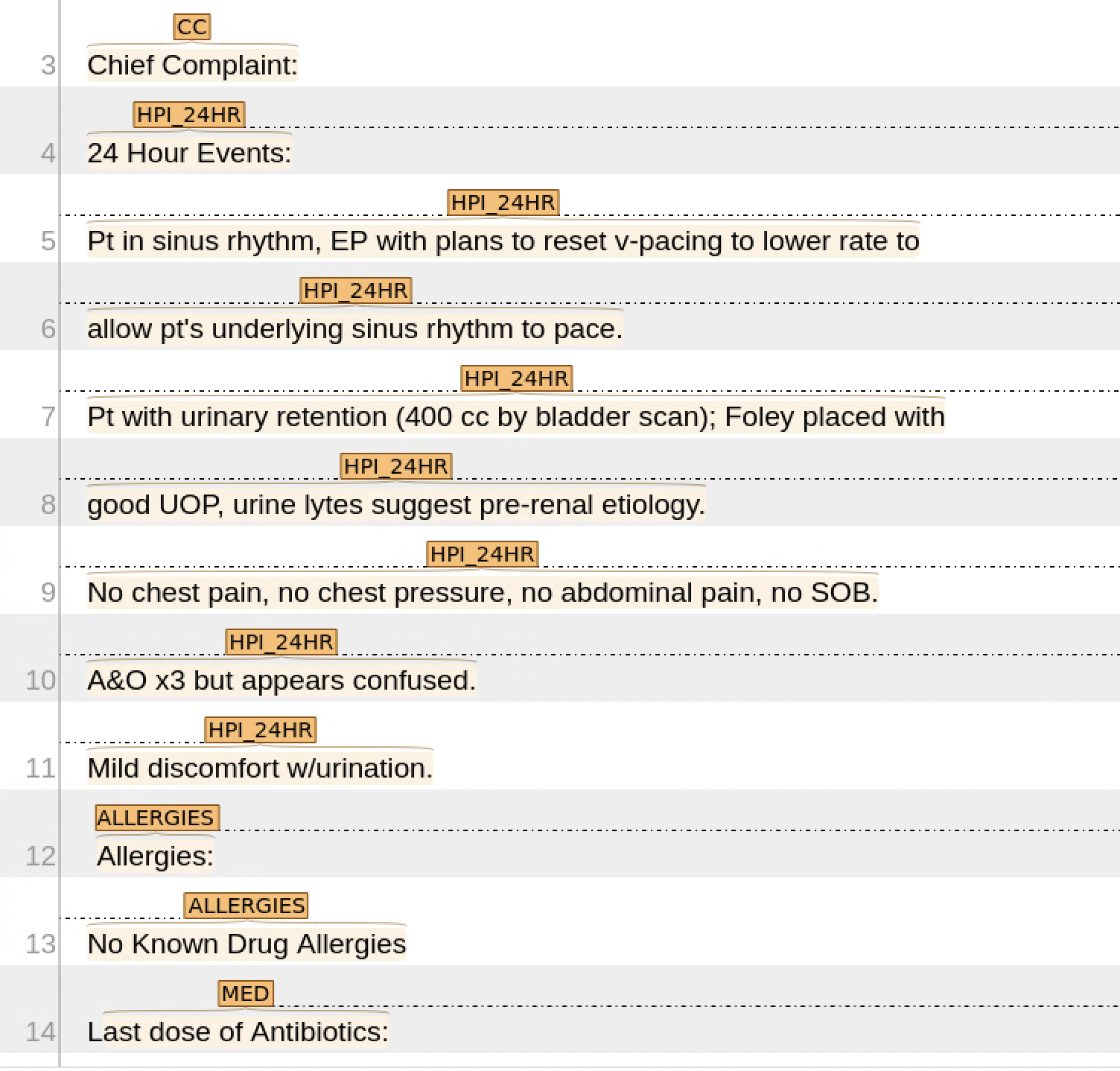}
    \vspace{-1.5pc}
    \caption{\small A screenshot of INCEpTION interface. The line numbers are indicated at left hand side. This piece of text are labelled with CC, HPI\_24HR, ALLERGIES, MED, indicating four sections.}
    \label{fig:section_tag}
\end{figure}

The annotations performed using INCEpTION were designed to provide labels and metadata in the daily progress notes extracted from MIMIC-III. MIMIC-III is an openly available dataset developed by the MIT Lab for Computational Physiology, comprising de-identified health data associated with approximately 60,000 intensive care hospitalizations. MIMIC includes demographics, vital signs, laboratory tests, medications, and over ten years of clinical notes collected across all the intensive care units at Beth Israel Hospital in Boston, Massachusetts, USA. To build the corpus of notes, a uniform random sample of progress notes across unique patient encounters were extracted from the NOTEVENTS table of MIMIC-III, which contained surgery, medicine, cardiovascular, neurology, and trauma daily progress notes. The annotations included an oversampling of medical progress notes because that is the largest service unit in the hospital.

\subsection{Note Type Selection}
The progress note types from MIMIC-III included a total of 84 note types (DESCRIPTION header) including the following: Physician Resident Note, Intensivist Note (SICU, CVICU, TICU), PGY1 Progress Note, PGY1/Attending Daily Progress Note MICU, MICU Resident/Attending Daily Progress Note. 
Other note types were excluded such as Nursing Progress Note and Social Worker Progress Note 
because these are not commonly structured in the SOAP  format. Further, history and physical notes (admission notes) and discharge summaries were excluded as these are not daily progress notes and preclude analytics at the hospital encounter level in a time-series manner.

\subsection{Ineligible Progress Notes}
After the sampling of notes was finalized, initial review by annotators was to assess appropriate listing of problems/diagnoses in the Assessment and Plan section. Progress notes with a Plan section that did not contain problems or diagnoses were excluded for annotation. In particular, many progress notes may be written with a systems-oriented Plan section which only lists organ systems with related details but without a diagnosis or problem identified (i.e., general systems-oriented note only include mentions such as ``Neurology", ``Cardiovascular", ``Renal", ``Hematology", etc.). These types of 
progress notes occurred in 741 (49\%) of the reviewed notes from the sampled corpus and were excluded for the following reasons: (1) they do not follow the problem-oriented view that is needed for clinical decision support; and (2) they are not the medical documentation best practices endorsed by medical school curriculum.

%% file: annotation.tex
\begin{figure*}[h]
    \centering
    \includegraphics[scale=0.33]{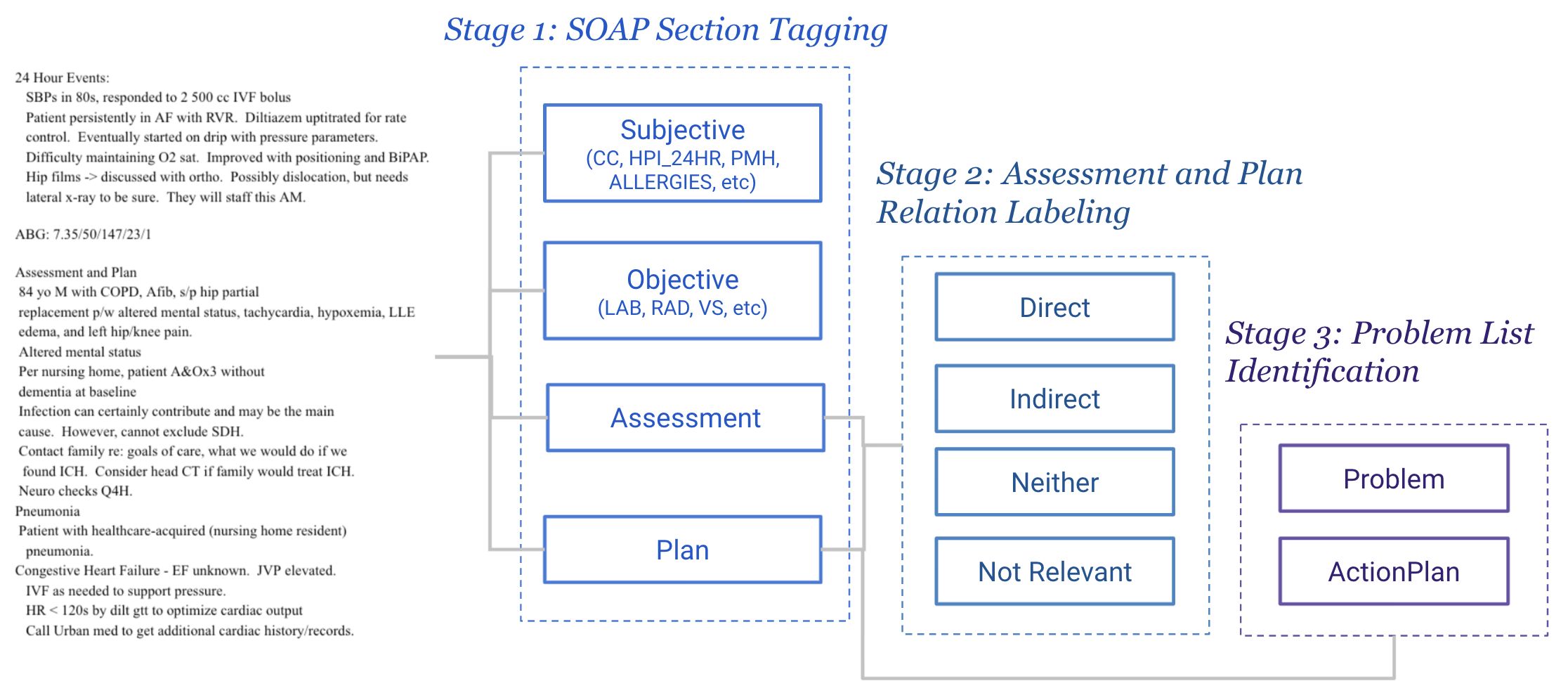}
    \vspace{-0.99pc}
    \caption{Annotation flow diagram for Progress Note Understanding. Dotted line boxes indicate different stages. Square text boxes are the labels produced at each stage.   }
    \label{fig:annotation}
\end{figure*}

\begin{table}
\small 
\centering
    \begin{tabular}{c|c|c} \hline 
        Sections & Attributes & SOAP Type  \\ \hline 
        Chief Complaint & CC & S \\ 
        History of Present Illness & HPI\_24HR & S \\ 
        Past Medical History & PMH & S \\ 
        Allergies & ALLERGIES & S \\
        Patient Surgical History & PSH & S \\ 
        Social History & SH & S \\ 
        Family History & FH & S \\ 
        Review of System & ROS & S \\ 
        Physical Exam & PE & O \\
        Medications & MED & O \\ 
        Laboratory & LAB & O \\
        Radiology & RAD & O \\ 
        Vital Sign & VS & O \\ 
        Assessment & PROBLEMLIST & A \\ 
        Plan & PLAN & P \\ 
        Addendum & ADDENDUM & OTHERS \\ 
        \hline 
    \end{tabular}
    \vspace{-0.4pc}
    \caption{Section headings in the progress notes, annotated tag attributes and corresponding SOAP types.}
    \label{tab:ann1}
\end{table}

\begin{figure*}
\small
\centering
    \begin{tabular}{l|l} \hline 
    Criterion & Label \\ 
    \hline 
       Assessment section includes a primary diagnosis/problem and it is mentioned in the Plan subsection.  &  \textsc{direct}\\
       Progress note includes a primary diagnosis/problem for hospitalization and it is mentioned in the Plan subsection. & \textsc{direct} \\ 
       Plan subsection contains a problem/diagnosis related to the primary signs/symptoms in the Assessment section. & \textsc{direct} \\ 
       Plan subsection contains complications/subsequent events or organ failure related to the primary diagnosis/ & \textsc{indirect}\\
       problem from the Assessment section. &  \\ 
       Plan subsection contains other listed diagnoses/problems from the overall Progress Note or in the Assessment  & \textsc{indirect} \\ 
       section that are not part of the primary diagnosis/problem. &  \\ 
       Plan subsection contains a diagnosis/problem that is not previously mentioned but closely related (i.e., same  & \textsc{indirect} \\
       organ system) to the primary diagnoses/problems mentioned in the overall Progress Note or Assessment section. &  \\ 
       None of the criteria for Directly Related or Indirectly Related are met but a diagnosis/problem or other & \textsc{neither} \\
       signs/symptoms are mentioned. &  \\ 
       Plan subsection does not include a diagnosis/problems OR signs/symptoms. & \textsc{not rel} \\ 
       \hline
    \end{tabular}
    \vspace{-0.9pc}
    \caption{Guidelines for annotating the four relations (\textsc{direct, indirect, neither, not relevant}) between Assessment and each subsection of Plan}
    \label{fig:ap}
\end{figure*}

We propose a hierarchical annotation schema consisting of three stages: \textit{SOAP Section Tagging} organizing all sections of the progress note into a SOAP category; \textit{Assessment and Plan Relation Labeling} specifying causal relations between symptoms and problems mentioned in the Assessment and diagnoses covered in each \textsc{Plan Subsection}; \textit{Problem List Identification} highlighting the final diagnoses. Every stage of the annotation builds on top of the previous annotation. Figure~\ref{fig:annotation} illustrates the flow of the annotation. The input to the flow is the entire SOAP progress note. The first stage of annotation, SOAP Section Tagging,  produces text segments labeled with different sections (example presented in Figure~\ref{fig:example_note}). Assessment and Plan sections are segmented at stage 1, and will be used as input to stage 2 Assessment and Plan Relation Labeling. The Plan section containing multiple subsections will be input to stage 3 for Problem List Identification, producing a list of diagnoses/problems.

\subsection{Progress Note Section Tagging}
The first stage of annotation is to segment the progress notes into sections, where each section belongs to a type of SOAP. Table~\ref{tab:ann1} presents a list of sections and headers that may appear in the progress notes and their corresponding \textsc{soap} types, following the definitions in~\cite{weed1964medical}. Given a progress note, the annotator will mark each line of text using one of the attributes, as shown in Figure~\ref{fig:section_tag}. Most of the sections start with a section header, indicating the lines below it falling into the same category of information until next section. When there is no section header, the annotator should mark the attributes by the content expressed in the lines (e.g. the line 14 \textit{Last dose of Antibiotics} belonging to \textsc{Medications}). We post-process the attribute labels and further categorize them as one of the SOAP type. A full description of the functionality of each section in \textsc{SOAP} notes could be found in~\cite{podder2020soap}.  

\subsection{Assessment and Plan Relation Labeling}
\begin{figure*}[h]
    \centering
    \includegraphics[width=\textwidth]{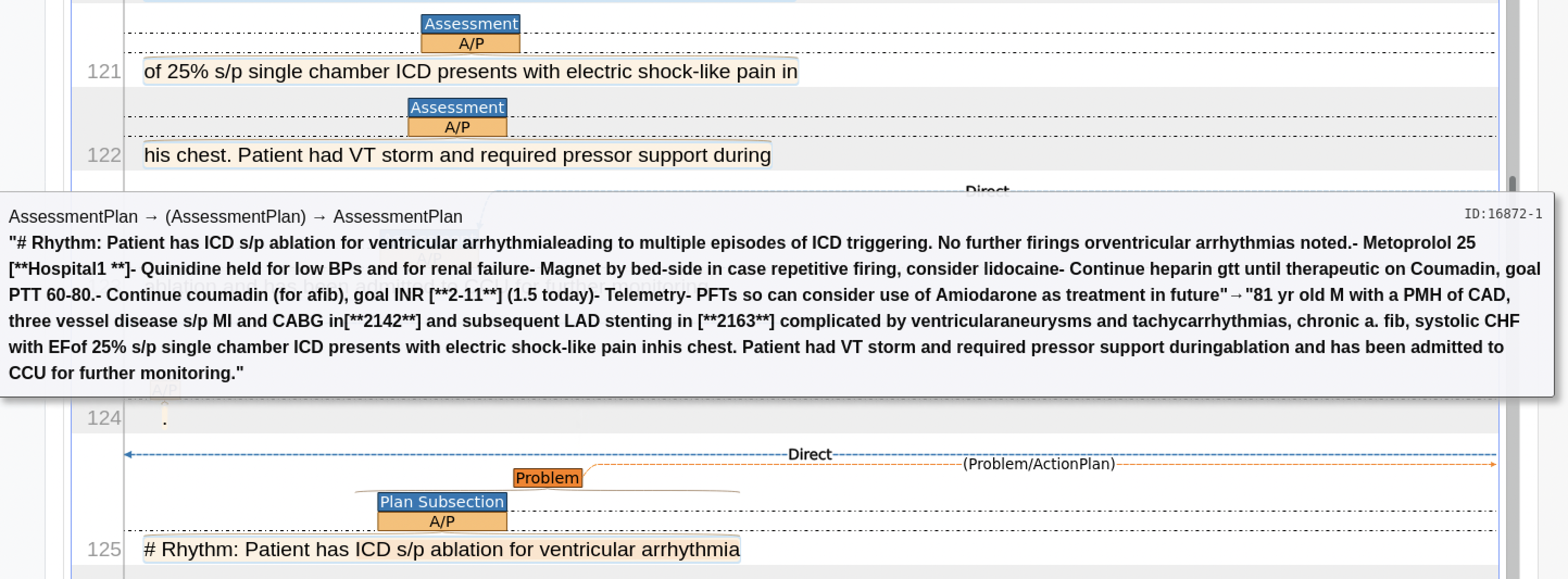}
    \vspace{-0.98pc}
    \caption{\small A screenshot of INCEpTION interface on \textsc{A\&P} Labeling. The \textsc{Plan Subsection} (line 125) contains a \textsc{Problem} \textit{ICD s\/p ablation for ventricular arrythmia}. A \textsc{direct} link connects the \textsc{Plan Subsection} and the \textsc{Assessment}, indicating that the problem is a major diagnosis. Once the link establishes, a text window pops up showing the text in the \textsc{Plan Subsection} (``\textit{Rhythm: Patient has ICD ... treatment in future}", and text in \textsc{Assessment} (``\textit{81 yr old M with a PMH ... further monitoring.}"), separated by a dash. }
    \label{fig:ap_labeling}
\end{figure*}




The Assessment and Plan (A\&P) sections are highlighted through the first stage of annotation. 
Recall that the Plan section contains multiple \textsc{Plan Subsection}, each of which lists a detailed plan for one specific diagnosis/problem. The second stage of annotation is to label \textsc{Plan Subsection} for its relation to the assessment: (1) directly related; (2) indirectly related; (3) not related; and (4) not relevant. These relations indicate whether a \textsc{Plan Subsection} addresses the primary diagnoses or problems related to the primary diagnosis (\textsc{direct}), an adverse event or consequence from the primary diagnosis or comorbidity mentioned in the Assessment (\textsc{indirect}), a problem or diagnosis not mentioned in the progress note (\textsc{neither}), or a Plan subsection without a problem or diagnosis listed (\textsc{not relevant}). Figure~\ref{fig:ap} shows the guidelines for marking the relations between the Assessment and each subsection of the Plan. Figure~\ref{fig:ap_labeling} presents an example of linking \textsc{Plan Subsection} with the Assessment using INCEpTION interface. 

\textsc{direct, indirect}, and \textsc{neither} relations all indicate that a diagnosis/problem is found in the progress note. When the label is \textsc{not relevant}, the plan subsection has no mention of a problem. Instead, it might describe quality improvement/administrative details such as physical therapy, occupational therapy, nutrition, prophylaxis (stress ulcer and gastric ulcer), or disposition. 

\begin{figure}[h]
    \centering
    \includegraphics[width=\columnwidth]{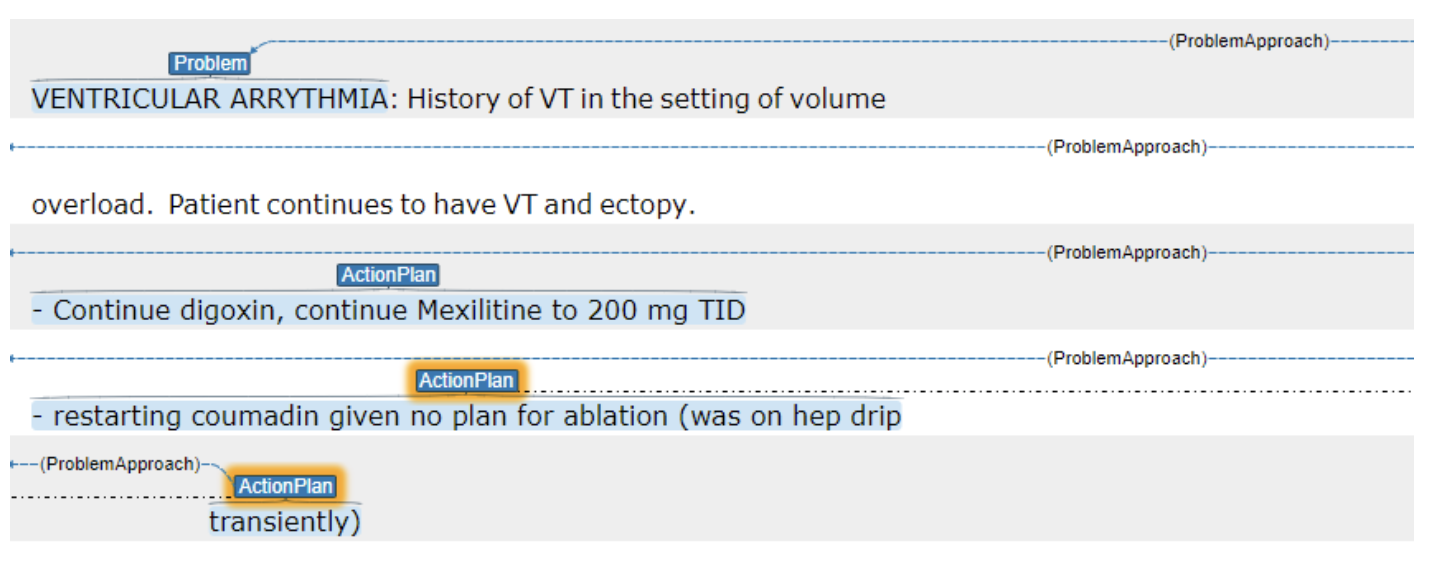}
    \vspace{-1.2pc}
    \caption{A Plan Subsection annotated with the \textsc{problem} (``Ventricular Arrythmia") and \textsc{ActionPlan}.}
    \label{fig:problem_action}
\end{figure}

\subsection{Problem List Identification}


The relevant plan subsections include a problem/diagnosis with an associated treatment or action plan, stating how the provider will address the problem. At the third stage of the annotation, the goal is to highlight the problems/diagnoses mentioned in the Plan subsections separately from the treatment or action plans for the day. In identifying problems/diagnoses, the annotators only labeled the text spans covering the problem in each Plan subsection, using the label \textsc{problem}. 
Once the problem/diagnosis was labelled then annotators labelled the accompanying \textsc{actionplan} for that \textsc{problem} and link the two attributes indicated by \textsc{problemapproach}. Figure~\ref{fig:problem_action} shows an example plan subsection where the \textsc{problem} is ``\textit{Ventricular Arrhythmia}", and the \textsc{actionplan} indicating the medical treatment for the ventricular arrhythmia.  

Problems sometimes include the syndrome (i.e., sepsis syndrome, acute respiratory distress syndrome) along with the underlying diagnosis (i.e, urinary tract infection, COVID-19 pneumonia). We ask the annotators to mark the entire span of text where the text contained both syndrome and/or diagnosis (e.g. ``\textit{sepsis likely due to urinary tract infection}", where \textit{sepsis} is the syndrome, and \textit{urinary tract infection} is the diagnosis). We include both syndrome and diagnosis because the same diagnosis may lead to different medical treatment pathways depending on the syndrome. For instance, antibiotics are needed for the \textit{urinary tract infection} and then there are resuscitation efforts and additional monitoring needed for the \textit{sepsis} syndrome. Phrases connecting between syndrome and diagnosis were included as part of the \textsc{problem} label, such as ``caused by” or ``likely due to” were included prior and after for diagnosis. These are strong linguistic cues showing the causal relations between syndrome and diagnosis, and reflecting a physicians' reasoning process. However, additional details about the diagnoses were not needed. For example, “Myocardial infarction involving the LAD s/p PCI” should be labelled only for the diagnosis/problem of “myocardial infarction”.  Another example, “SVT with hypotension: This occurred in setting of having RA/RV manipulated with wire” should be labelled as just “SVT with hypotension”. These were descriptions of characteristics of the diagnosis. 

\subsection{Annotator Training}

The annotation guidelines and rules were initially developed and trialed by two physicians with board certifications in critical care medicine, pulmonary medicine, and clinical informatics. The physicians practice in the same field as the authors of the source notes. They are also experts in clinical research informatics with an extensive research track in machine learning and natural language processing, one of them serves as the mentor and adjudicator for the trained annotators. Two medical students were recruited and had received training in their medical school curriculum in medical history taking and documentation (including SOAP format), anatomy, pathophysiology, and pharmacology.  An additional three week period with orientation and training was provided by one of the critical care physicians to the annotators. Each annotator met an inter-rater reliability with a kappa score of $>0.80$ with the adjudicator prior to independent review. The annotators augmented their medical knowledge with a subscription and access to a large medical reference library, UpToDate®\footnote{\url{https://www.wolterskluwer.com/en/solutions/uptodate}}. The adjudicator performed audits of the charts after approximately 200 were completed and if the inter-rater reliability fell below the 0.80 threshold then cases with disagreements were reviewed to consensus and the annotator was re-trained again until threshold kappa agreement was met.  

%% file: results.tex
In total, two annotators labelled 768 progress notes with 28,945 labels from all annotation stages.  We further split the corpus into train/dev/test, resulting in 608, 76 and 87 notes, respectively. All annotations were stored as XML files. We reported statistics of the labels for each stage.

\begin{table}
\small 
\centering
    \begin{tabular}{c|c|c|c} \hline 
         Attributes & \#Labels & Avg \#Labels & Avg   \\ 
         & & Per Note & Length \\ \hline 
         CC & 750 & 0.98 & 12.70\\ 
         HPI\_24HR & 807 & 1.51 & 94.37 \\ 
         PMH & 731 & 0.95 & 7.42\\ 
         ALLERGIES & 764 & 0.99 & 17.24\\
         PSH & 12 & 0.02 & 33.91 \\ 
         SH & 15 & 0.02 & 15.87 \\ 
         FH & 711 & 0.93 & 5.05\\ 
         ROS & 740 & 0.96 & 21.61 \\ 
         PE & 789 & 1.03 & 131.64\\
        MED & 823 & 1.07 & 76.31\\ 
         LAB & 3083 & 4.01 & 55.59   \\
         RAD & 1395 & 1.82 & 38.75\\ 
         VS & 974 & 1.27 & 159.92 \\ 
         A\&P & 787 & 1.02 & 577.29 \\
         Addendum & 727 & 0.95 & 47.62\\ 
        \hline 
    \end{tabular}
     \vspace{-0.4pc}
    \caption{Total and average number of section tags, and average length for each section counting by tokens.  }
    \label{tab:section_count}
\end{table}

\begin{figure}
    \centering
    \includegraphics[width=\columnwidth]{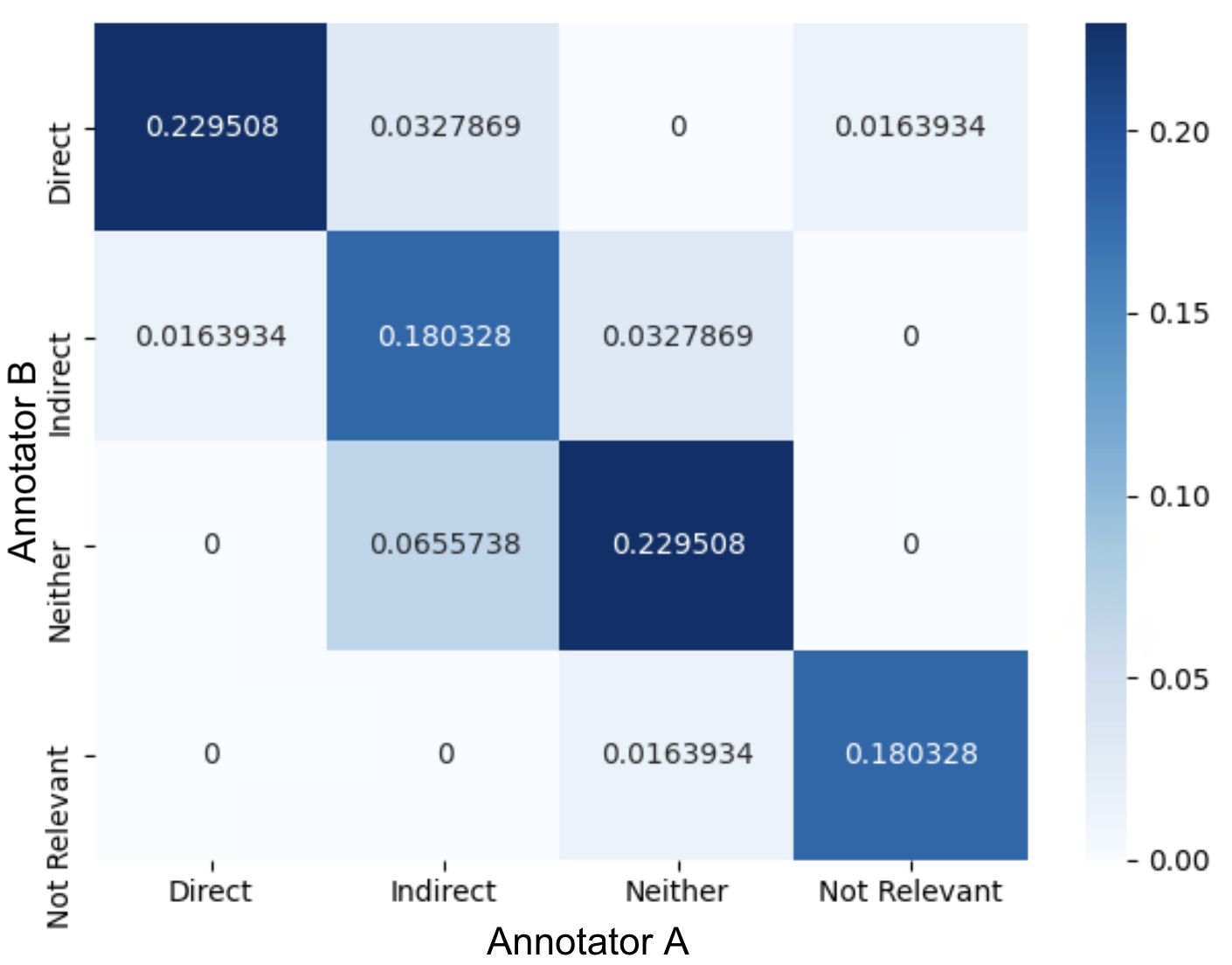}
    \vspace{-0.99pc}
    \caption{Confusion matrix on the \textsc{AP} relation labels between two annotators.}
    \label{fig:inter_ann}
\end{figure}

\paragraph{Statistics for Section Tagging} We collected 3,790, 6,090, 787 and 2,742 labels for SOAP and \textsc{Others}, respectively. Table~\ref{tab:section_count} presents the statistics of labels broken down by each section attribute, showing the total and average number of section tagged and average length of sections by tokens. Laboratory (LAB) was the most frequent section type across our sampled progress notes, with 4 sections per note on average. Often, more than one lab test was required to provide evidences for certain diseases, and every lab tested was included in the progress note as an individual section. Assessment and Plan sections were the longest with 577.29 tokens per note, much more than any other sections. 
Recall that Assessment and Plan viewed as the most important piece in SOAP note, summarizing the evidence from other sections and listed diagnoses with treatment plans. Physicians tend to spend most of their time reading the Assessment and Plan sections and less than 10 percent of the content in physicians' verbal handoffs were found outside the A/P section~\cite{brown2014physicians}. 

\begin{table}
\small 
\centering
    \begin{tabular}{c|c|c|c|c} \hline 
    Count & \textsc{direct} & \textsc{indirect} & \textsc{neither} & \textsc{not rel} \\ \hline 
    Total & 1404 & 1599 & 1913 & 1018 \\ 
    Per Note & 1.83 & 2.09 & 2.49 & 1.32 \\ 
    \hline 
    \end{tabular}
    \vspace{-0.8pc}
    \caption{Total and average number of relation labels.  }
    \label{tab:label_count}
    \end{table}
\paragraph{Statistics of A\&P Relation Labeling}  Recall that A\&P relations were labelled between every pair of \textsc{Plan Subsection} and the Assessment. Across the sample set of progress notes, we had 7.73 plan subsections per note on average. Table~\ref{tab:label_count} summarizes the count of labels on all annotated notes and the average per note. The distribution across four relations were relatively balanced, with \textsc{neither} being the most frequently labelled and \textsc{not relevant} being the least frequently labelled.  

\paragraph{Statistics of Problem Lists} We collected 4,843 and 4,759 labels for \textsc{problem} and \textsc{actionplan}, respectively. For every note, the average numbers of \textsc{problem} and \textsc{actionplan} were both approximately 6. Recall that the \textsc{problem} was highlighted after the annotation for \textsc{a\/p} Relation Labeling, we were able to count the problems that were labeled as \textsc{direct} and \textsc{indirect}, which was 2,866 in total. We post-processed the annotation such that for every assessment, there was a list of direct problems and a list of indirect problems. Every list could be regarded as a short summary, with distinct problems delimited by semi-colons. In total we collect 743 and 619 summaries for direct problems and indirect problems, respectively.

\paragraph{Inter-Annotator Agreement} We measured Cohen's Kappa on the \textsc{AP} relation labeling task, as it was deemed the most difficult by the annotators because it was the only task that required clinical reasoning and medical knowledge. The two annotators achieved a Cohen's Kappa of 0.74 on 10 randomly sampled notes, which represented good quality given the complexity of the task. Figure~\ref{fig:inter_ann} presents the agreement and disagreement between the two annotators on four relations. Most of the disagreement occurred between \textsc{indirect} and \textsc{neither}, taking 9.83\% of all labels, exposing the difficulties in deciding whether a problem is due to a subsequent event of the main diagnosis (\textsc{indirect}) or a separate disease altogether (\textsc{neither}). Nonetheless, the percentage of labels agreed by both annotators was 81.9\%.

%% file: task1.tex
\begin{figure*}
    \centering
    \includegraphics[scale=0.4]{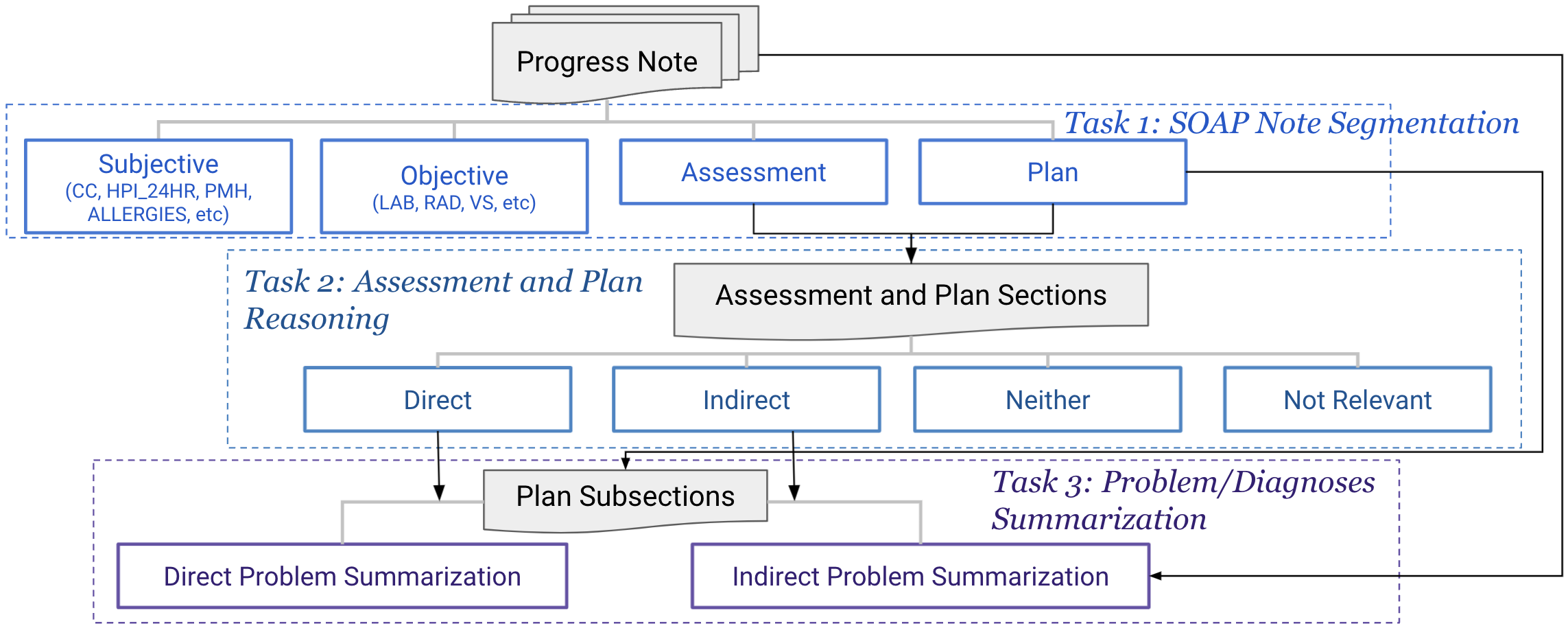}
    \vspace{-0.9pc}
    \caption{\small The \textit{Progress Note Understanding} suite is consisted of three tasks, each of which corresponds to one annotation stage. Task 3, \textit{Problem/Diagnoses Summarization}, has two subtasks, \textit{Direct Problem Summarization} and \textit{Indirect Problem Summarization}, using the problems identified in annotation stage 2 and 3. }
    \label{fig:suite}
\end{figure*}

In this section, we further propose the suite of clinical NLP tasks: \textit{Progress Note Understanding}. The suite of tasks was developed from the annotation and designed to train and evaluate models for clinical text understanding in future work. 
The suite of tasks was set up in the same stream as the annotation, with each task corresponding to an annotation stage and targeted at a different NLP problem. Figure~\ref{fig:suite} presents a diagram of the suite set up. The level of NLP difficulty in moving from Task 1 to Task 3 were intended to increase from classification to classification with medical knowledge (NLI) to text generation. We consider the SOAP Note Section Segmentation as the easiest task that only requires text understanding, and the Diagnoses Summarization to be the hardest as it investigates clinical text understanding, clinical reasoning and text generation. 


\subsection{Task 1: SOAP Note Segmentation}

The suite of tasks started with a clinical text understanding task where the goal was to segment the entire SOAP note into topic-relevant sections. We formulated this task as labeling each line of the daily progress note by its \textsc{soap} type based on the information contained in the line. This task helped to train and evaluate models that automatically understand the SOAP note structure and topic-level contents, and highlights the problem-oriented sections. We considered this task as a fundamental and essential step for NLP system pipelines for clinical note processing, as previously shown in~\cite{krishna2021generating,cillessen2012modeling,mowery2012building}. Previous work also focused on segmenting SOAP notes~\cite{mowery2012building,ganesan2014general}. \newcite{mowery2012building} formulated the task as sentence labeling using 50 emergency department reports. Our proposed task is similar to \newcite{ganesan2014general} where the task is on line-basis instead of sentence, but our work is a larger corpus and is focused on daily progress notes.  Lastly, we plan to make our corpus available unlike prior work. Evaluation will be measured using the F1 score and Accuracy as in previous work~\cite{mowery2012building}.

\subsection{Task 2: Assessment and Plan Reasoning}
Recall that the Assessment and Plan sections contain the most useful information for providers needing to know what is the condition of a patient. Providers spend the most time viewing these sections, where clinical reasoning occurs. Providers conclude the diagnoses/problems from the evidences presented in \textsc{S} and \textsc{O} sections and the symptoms described in \textsc{A} section, then infer the treatment plan for each diagnosis for that day. It is a process of associating the patients' current condition with clinical problems, and assessing the solutions with medical knowledge. Modeling such understanding and reasoning process greatly benefits the downstream applications that aim at improving efficiency of bedside care and reducing medical errors. We propose task 2, Assessment and Plan Reasoning, a relation classification task that builds on top of the stage 2 annotation. Models trained and evaluated on this task should predict one of four relations. We plan to use F1 score as the evaluation metric for this task.\footnote{The task is part of 2022 National NLP Clinical Challenge (N2C2): \url{https://n2c2.dbmi.hms.harvard.edu/2022-track-3}. } 

\subsection{Task 3: Problem/Diagnoses Summarization}
Automatically generating a set of diagnoses/problems in a progress note could help providers quickly and efficiently understand and document a patient's condition, and ultimately reduce effort in document review and augment care during time-sensitive hospital events. \newcite{Devarakonda2017AutomatedPL} demonstrated that a system that automatically predicts diagnoses captures more problems than human clinicians through a pilot study. They built a two-level classifier that first predicted major category of conditions and then a subsequent prediction on the fine-grained level of problems. \newcite{liang2019novel} formulated the task as extractive summarization on a disease-specific progress notes. To enrich the research on diagnosis summarization, we propose \textit{Problem/Diagnoses Summarization} with labels created from annotation stage 2 and 3. Specifically, the task consisted of two subtasks: Direct Problem Summarization and Indirect Problem Summarization. 
The goal of Direct Problem Summarization is to take the assessment section as input, and predict a list of primary diagnoses. For Indirect Problem Summarization, the goal is to predict a list of adverse events and subsequent diagnoses given the entire progress note. Recall that in Figure~\ref{fig:ap_labeling}, indirect problems included diagnoses/problems that were not part of the Assessment or primary diagnoses, which required input from other sections in the progress notes (e.g. LAB, VS) and would make the task harder than Direct Problem Summarization. Different from \cite{liang2019novel}, we defined both subtasks as abstractive summarization, since our data covered a larger breadth of diseases/problems and they were not always explicitly mentioned in the progress note. In some instances, the signs or symptoms of a disease were listed, relevant objective data that pointed to a diagnosis, or synonyms and acronyms of the disease or problem were listed in other parts of the note. Recall that the text spans labelled by \textsc{Problem} were concatenated through semi-colons. Figure~\ref{fig:direct_summ} includes an example Direct Problem Summary with an assessment it derived from.

\begin{figure}
    \centering
    \small
    \begin{tabular}{l} 
    \hline 
    Assessment \\ 
    \hline 
         80 yo female with pmh of htn and DM admitted with an \\ 
         NSTEMI transferred to the CCU after an attempted cath due  \\
         to a hypertensive emergency secondary to anxiety and \\
          agitation.  \\ 
 \hline 
   Direct Problem Summary \\ \hline 
        multivessel CAD; Hypertensive emergency; Change in  \\
        mental status/agitation; \\ 
 \hline 
    \end{tabular}
    \vspace{-0.8pc}
    \caption{ \small An example Direct Problem Summary given an Assessment section. The diagnosis ``Multivessel CAD" is inferred from ``DM admitted with an NSTEMI transfered to CCU"; ``hypertensive emergency" is mentioned directly in the assessment, and ``Change in mental status/agitation" refers to ``anxiety and agitation". }
    \label{fig:direct_summ}
\end{figure}


For both summarization subtasks, we plan to follow previous work on EHR summarization~\cite{macavaney2019ontology,adams2021s} and use the standard summarization metrics like ROUGE~\cite{lin2004rouge}. A diagnosis could be expressed with different lexicons, (e.g. ``CAD" and ``Coronaries"), hence we also consider post-processing the summaries with the National Library of Medicine's Unified Medical Language System (UMLS)~\cite{bodenreider2004unified} by mapping the clinical terms to the UMLS metathesaurus of concept unique identifiers (CUIs), and measuring the F1 score for clinical accuracy. \newcite{liang2019novel} also used this method as their main evaluation.